\documentclass[runningheads]{llncs}
\usepackage[T1]{fontenc}
\usepackage{amsmath}
\usepackage{graphicx}
\usepackage{appendix}
\usepackage{hyperref}
\usepackage[square,numbers]{natbib}
\usepackage{multirow}
\usepackage{rotate}
\usepackage{array}
\usepackage{array,calc} % For defining column widths
\usepackage{adjustbox} % For better text alignment inside rotated columns

\usepackage{color}

\DeclareMathOperator*{\argmin}{arg\,min}
\newcommand{\etal}{\textit{et al.\ }}

\begin{document}
\title{Diffusion Counterfactuals for Image Regressors}
\author{Trung Duc Ha\orcidID{0009-0001-2461-6833} \and
Sidney Bender\orcidID{0009-0007-1727-8577}}
% \author{Trung Duc Ha \and Sidney Bender}
%
\authorrunning{Trung Duc Ha, Sidney Bender}
\institute{Technische Universität Berlin, Straße des 17. Juni 135,
10623 Berlin, Germany
% \url{https://web.ml.tu-berlin.de}
\\
\email{\{t.ha,s.bender\}@tu-berlin.de}}
\maketitle              % typeset the header of the contribution
\begin{abstract}
% The abstract should briefly summarize the contents of the paper in

% 1. **Abstract**
%    - Brief summary of the paper’s goal
%    - Key methodology and findings
%    - Main contributions in a concise manner

Counterfactual explanations have been successfully applied to create human interpretable explanations for various black-box models. They are handy for tasks in the image domain, where the quality of the explanations benefits from recent advances in generative models.
Although counterfactual explanations have been widely applied to classification models, their application to regression tasks remains underexplored.
We present two methods to create counterfactual explanations for image regression tasks using diffusion-based generative models to address challenges in sparsity and quality: 1) one based on a Denoising Diffusion Probabilistic Model that operates directly in pixel-space and 2) another based on a Diffusion Autoencoder operating in latent space.
Both produce realistic, semantic, and smooth counterfactuals on CelebA-HQ and a synthetic data set, providing easily interpretable insights into the decision-making process of the regression model and reveal spurious correlations. 
We find that for regression counterfactuals, changes in features depend on the region of the predicted value. Large semantic changes are needed for significant changes in predicted values, making it harder to find sparse counterfactuals than with classifiers. 
Moreover, pixel space counterfactuals are more sparse while latent space counterfactuals are of higher quality and allow bigger semantic changes.

\end{abstract}

% ---------------- MAIN CONTENT ----------------
% 1. Introduction
\section{Introduction}
% ---------------- CONTENT ----------------
% Introduction
%    - Background on explainable AI and counterfactual explanations
%    - Relationship between counterfactual explanations and adversarial attacks
%    - Motivation for the study
%    - Key contributions of the paper
% ---------------- CONTENT ----------------

% Discuss the importance of explainability in decision-making processes.
% Describe how neural networks are often seen as "black boxes."
% Introduce the current state of explainability research, focusing on post-hoc methods, with a specific focus on attribution and counterfactual methods.
% Mention the domain we are going to do this in (images) and current available methods.
% Explainable Artificial Intelligence (XAI) is a set of methods and algorithms to enable these properties for predictions. 
Interpreting the results of deep learning (DL) models is still challenging due to the black-box nature of the predictions. In particular, the model might be using features that indicate good performance while completely relying on spurious correlations, commonly referred to as the ``clever hans effect'' \cite{andersFindingRemovingClever2022,  lapuschkinUnmaskingCleverHans2019, samekExplainingDeepNeural2021,samekExplainableArtificialIntelligence2019}. In addition, understanding these results is critical for ethical and safety reasons in fields such as healthcare, credit scoring, and more. DL models must guarantee trustability, transparency, and fairness to be deployed in these areas \cite{dasOpportunitiesChallengesExplainable2020}.

This paper will focus on the post hoc explanation method called \emph{counterfactual explanations}. The goal is to provide a ``what-if`` scenario of an alternate input that changes the prediction of the model while constraining the changes to be minimal and semantically meaningful \cite{guidottiCounterfactualExplanationsHow2022}. This creates counterexamples that are more likely to be intuitive and easily interpretable \cite{vermaCounterfactualExplanationsMachine2020}. Thus, users may arrive at natural conclusions such as ``had I only changed $X$, the system would have predicted $Y'$ rather than $Y$.''

Particular interest has been given to image counterfactuals using generative models, guided by gradients toward altering the prediction. These methods can be divided into two categories: Using latent representations \cite{dombrowskiDiffeomorphicCounterfactualsGenerative2024, rodriguezTrivialCounterfactualExplanations2021} or derivatives of diffusion models \cite{atadCounterfactualExplanationsMedical2024, jeanneretDiffusionModelsCounterfactual2022, jeanneretAdversarialCounterfactualVisual2023,  jeanneretTextImageModelsCounterfactual2023} to create changes along the data manifold. The effectiveness of these methods hinges on image quality, leading to increased adoption of diffusion models.

% Image Classificaiton -> Regression
Most CE methods focus on image classification, as in the previous work mentioned. However, many important image tasks involve regression, especially in pathology \cite{hessePrototypeLearningExplainable2024, cohenGifsplanationLatentShift2021, singlaExplainingBlackboxSmoothly2023,vanderveldenVolumetricBreastDensity2020}. 
 % Intro to XAIR
Explainable AI for Regression (XAIR) methods have emerged to form a theoretical foundation \cite{khanExplainableRegressionFramework2022,letzgusExplainableAIFramework2024, letzgusXpertAIUncoveringModel2024, letzgusExplainableAIRegression2022}. Specifically, selecting an appropriate reference point is essential for producing the right explanations. In the context of XAIR, this is referred to as the \emph{reference value} \cite{letzgusExplainableAIRegression2022}. It serves as an important parameter for the context of a regression explanation, e.g. ``what would this person look like if they were 20 or 80 years old?''. This approach allows for more targeted explanations, distinguishing it from classifier explanations. 

Some counterfactual approaches for image regression exist \cite{ atadCounterfactualExplanationsMedical2024,dombrowskiDiffeomorphicCounterfactualsGenerative2024}. However, these approaches either use low-resolution generative models or do not use pre-trained regressors. Moreover, they do not consider reference values. This gap forms the foundation of the contributions of this paper.
% Our Solution
% TODO: Need to cover the challenges of regression: Sparsity vs traversing the error surface
Image regression counterfactuals present unique challenges: How do we achieve minimal, meaningful, and realistic changes while applying them effectively to regression? Moreover, defining minimality through pixel footprint may decrease interpretability \cite{delaneyCounterfactualExplanationsMisclassified2022}.

To address these issues, our contributions are twofold:
(1) We present two methods to create counterfactual
explanations for image regression tasks using diffusion-based generative
models. We adapt Adversarial Counterfactual Explanations (ACE) \cite{jeanneretAdversarialCounterfactualVisual2023} that operate directly in pixel space and the Diffusion Autoencoder (Diff-AE) \cite{preechakulDiffusionAutoencodersMeaningful2022} that operates in latent space.
(2) Regression-specific adaptations that allow us to produce counterfactuals with higher granularity, inspecting specific regions of the predicted values.

% Key results and summary of findings
Our experiments on CelebA-HQ and a synthetic dataset demonstrate that both methods produce realistic, semantic, and smooth counterfactuals. We can reveal spurious correlations and observe that feature changes depend on the prediction region, with larger semantic alterations required for significant value shifts. Furthermore, we find a trade-off between sparsity and quality, with pixel-space changes offering greater sparsity, and latent-space edits providing higher quality and semantic flexibility.

% Outline of paper structure
The remainder of this paper is organized as follows. Section \ref{sec:related_work} reviews the relevant literature on counterfactual explanations and existing regression explanation techniques. Section \ref{sec:methodology} includes an overview of diffusion models and their derivatives and elaborates on our two novel approaches. Section \ref{sec:experiments} details our experimental setup with datasets, evaluation methods, and their results. It also includes implementation details, an ablation study, and regression-specific explanation analysis with spurious correlation identification. Finally, Section \ref{sec:conclusion} concludes the paper with a summary of our findings and a discussion of potential future research directions.

\section{Related Work}\label{sec:related_work}
% ---------------- CONTENT ----------------
% Related Work
%    - Overview of explainable AI, including post-hoc vs. ad-hoc methods
%    - Counterfactual explanations (existing methods and challenges)
%    - Adversarial attacks and their relation to counterfactual explanations
%    - Prior use of diffusion models in explainability
%    - Prior work of regression based counterfactuals
%    - regression problems in general
%    - Mention DiffeoCF as well?
% ---------------- CONTENT ----------------

We contextualize our work by reviewing key studies on XAI, counterfactual explanations, and regression explanations relevant to our approach. Furthermore, we highlight the position of our work within the limited research of image regression counterfactuals.

% \subsection{Explainable Machine Learning}
\subsection{Counterfactual Explanations for XAI}

% \subsection{Counterfactual Explanations}
% \begin{itemize}
%     \item approaches pixel space, latent space
%     \item mention \cite{dombrowski2021diffeomorphic}
% \end{itemize}

XAI methods generally align themselves by a mixture of the following properties \cite{dasOpportunitiesChallengesExplainable2020, guidottiCounterfactualExplanationsHow2022, molnar2022, samekExplainingDeepNeural2021}: 
The scope of a method describes whether it is applicable to the \textit{global} behavior of the model or to a single \textit{local} data point. 
% In addition, methodology is distinguished between the application domain: \textit{pertubation-based} methods iteratively change features in the input data to attribute them their importance, while \textit{backpropagation-based} methods rely on the computations of gradients of a neural network.
Its usage categorizes it as interpretable by design (\textit{intrinsic}) or architecture independent, applying to outputs of pre-trained models (\textit{post-hoc}).

Local post-hoc methods are highly useful in areas such as healthcare and finance \cite{dasOpportunitiesChallengesExplainable2020} as they can be directly applied to deployed blackbox models to explain highly critical decisions. Although common methods belonging to this class \cite{bachPixelwiseExplanationsNonlinear2015, baehrensHowExplainIndividual2010, lundbergUnifiedApproachInterpreting2017, ribeiroWhyShouldTrust2016, sundararajanAxiomaticAttributionDeep2017, zeilerVisualizingUnderstandingConvolutional2014, zhouLearningDeepFeatures2016} visualize the importance of features of a particular prediction, they do not suggest \textit{actionable} insights. Counterfactual explanations (CE) \cite{wachterCounterfactualExplanationsOpening2017} have been developed to solve this gap. CEs create a counterexample of an input that alters the model's prediction, restricting the changes to be minimal and semantically meaningful. They reveal the most sensitive features in an intuitively interpretable manner. This allows users to understand patterns that influence the model's decision-making and identify potential vulnerabilities or biases. Moreover, CEs are suitable for finding and removing spurious correlations \cite{benderFixingCleverHansPredictors2023a, jeanneretAdversarialCounterfactualVisual2023}.

In image applications, generated counterfactuals must be realistic and lie on the data manifold. Several approaches exist; DiVE \cite{rodriguezTrivialCounterfactualExplanations2021} employed a Variational Autoencoders \cite{kingmaAutoEncodingVariationalBayes2022} to perform modifications in its latent space, restricting it to the data manifold. Diffeomorphic Counterfactuals \cite{dombrowskiDiffeomorphicCounterfactualsGenerative2024} also followed this approach, additionally using Generative Adversarial Networks (GAN) to do this task, supplementing it with strong theoretical guarantees. Similarly, STEEX \cite{jacobSTEEXSteeringCounterfactual2022} used GANs in conjunction with a semantic map to restrict edits. However, these methods are limited by their generative models, which struggle to create high-resolution images.

Diffusion models \cite{hoDenoisingDiffusionProbabilistic2020} addressed this problem, leading to adoption for generating counterfactual images. DiME \cite{jeanneretDiffusionModelsCounterfactual2022} uses diffusion models to partially noise the image and then guide the denoising process using a classifier. Similarly, ACE \cite{jeanneretAdversarialCounterfactualVisual2023} starts by only partially noising the image. However, they based their method on adversarial attacks \cite{szegedyIntriguingPropertiesNeural2014} using the diffusion process to filter out non-semantic components. In addition, the authors employ RePaint \cite{lugmayrRePaintInpaintingUsing2022} for further refinement.

Counterfactual generation shares goals with semantic editing: making meaningful changes while maintaining overall structure and coherence. Methods involve GANs \cite{cherepkovNavigatingGANParameter2021, lingEditGANHighPrecisionSemantic2021} and more recently diffusion models \cite{couaironDiffEditDiffusionbasedSemantic2022, huberman-spiegelglasEditFriendlyDDPM2024, kwonDiffusionModelsAlready2023}. We highlight the Diffusion Autoencoder (Diff-AE) \cite{preechakulDiffusionAutoencodersMeaningful2022}, which encodes high-resolution images into an editable semantic latent space while accurately reconstructing them. For interpretability, semantic edits are favored by humans, compared to minimal edits enforced by counterfactual search \cite{delaneyCounterfactualExplanationsMisclassified2022}.

% latent editing (GANS, Diff-AE)
% Highlight DiffeoCF as a major influence

\subsection{Regression Explanations}

% General applications
Several works apply classification-based attribution methods to regression problems either directly   \cite{khanExplainableRegressionFramework2022, vanderveldenVolumetricBreastDensity2020} or by first converting the regression task into multi-class classification \cite{binderMorphologicalMolecularBreast2021, lapuschkinUnderstandingComparingDeep2017}.
% Contrast to out of the box application
Letzgus \etal \cite{letzgusExplainableAIRegression2022} suggest that regression explanation methods need adaptation from classification. Since regression involves real-valued predictions, specifying a \textit{reference value} for the explanation method is crucial to align it with the intention of the user. For example, ``why an item is currently valued at 1200 dollars compared to its usual 1000 dollars price'' \cite{letzgusExplainableAIRegression2022}. Therefore, the user can provide context and precisely target the explanation. In addition, the reference value can be integrated into the measurement unit of the regression problem. Applications of this methodology include \cite{letzgusXpertAIUncoveringModel2024} and \cite{letzgusExplainableAIFramework2024}.

Regression counterfactual explanations have also been employed for regression tasks on structured \cite{spoonerCounterfactualExplanationsArbitrary2021} and multivariate time series \cite{ramanWhyDidThis2023} data. 
Applied to images, Dombrowski \etal \cite{dombrowskiDiffeomorphicCounterfactualsGenerative2024} generate counterfactuals on the Mall dataset \cite{chenFeatureMiningLocalised2012}, only minimizing/maximizing the predicted number of pedestrians, and do not cover interpolation between explanations. 

Several works cover counterfactuals for ordinal regression on pathological images \cite{cohenGifsplanationLatentShift2021, klauschenExplainableArtificialIntelligence2024, singlaExplainingBlackboxSmoothly2023}. We highlight the work of Atad \etal \cite{atadCounterfactualExplanationsMedical2024}, who apply this to diffusion models by using the Diff-AE and a regressor predicting in the latent space.

This distinction discerns our contribution: using diffusion models to generate high-quality counterfactual images, and explaining a regressor operating in pixel space while considering reference values.

% 3. Main Method
\newcommand{\lossReg}{L_\text{reg}}
\newcommand{\lossClass}{L_{\text{class}}}
\newcommand{\confidence}{c}
\newcommand{\filterf}{F}
\newcommand{\zsem}{\mathbf z_{\text{sem}}}
\newcommand{\yref}{\tilde y}
\newcommand{\ypred}{\hat y}
\newcommand{\enc}{\operatorname{Enc}_{\phi}}
\newcommand{\nnDae}{\mathbf f_\theta}
\newcommand{\nnReg}{\mathbf r_\psi}
\newcommand{\xIn}{\mathbf{x}}
\newcommand{\xRec}{\hat{\mathbf{x}}}
\newcommand{\xCf}{\mathbf{x}'}

\section{Methodology}\label{sec:methodology}
% ---------------- CONTENT ----------------
% Methodology
%    - Problem Definition: Mathematical formulation of the problem
%      - Loss functions and optimization
%      - Details on how adversarial attacks are leveraged for counterfactual generation
%    - Proposed Approach:
%      - Explanation of the novel method 
%      - Steps of the approach, including adversarial attack generation and other procedures
%      - Use of Denoising Diffusion Probabilistic Models (DDPMs) in ACE and Diff-AE
%      - Algorithm?
% ---------------- CONTENT ----------------

In this section, we discuss our methods for the generation of regression counterfactuals. We first briefly summarize and review Denoising Diffusion Probabilistic Models (DDPM) and the relevant derivatives, the Denoising Diffusion Implicit Models (DDIM), and the Diffusion Autoencoder (Diff-AE). Lastly, we suggest two novel approaches to create regression counterfactuals: adaptations for ACE to form Adversarial Counterfactual Regression Explanations (AC-RE) and Diff-AE Regression Explanations (Diff-AE-RE).

% ---------------- DRAFT ----------------

\subsection{Diffusion Models and their Derivatives}\label{sec:diffusion}

Denoising Diffusion Probabilistic Models (DDPMs) \cite{hoDenoisingDiffusionProbabilistic2020} belong to a family of models that learn to model a data distribution through a forward and reverse noising process. The forward process gradually adds Gaussian noise to a data sample $\mathbf x_0$ over timesteps $1 \leq t \leq T$, creating a sequence of increasingly noisy samples. It is possible to directly infer the noised state $\mathbf x_t$ from any timestep $t$ as 
\begin{equation}\label{eq:ddpm_f}
\mathbf{x}_t = \sqrt{\alpha_t} \mathbf{x}_0 + \sqrt{1 - \alpha_t} \epsilon, \quad \epsilon \sim \mathcal{N}(\mathbf{0}, \mathbf{I}).
\end{equation}
where $\alpha_t$ is the time-dependent noise schedule and $\epsilon$ is a noise variable.

A network then learns to reverse this noising process through a Markovian process, starting from $t=T$ and finally reaching $\mathbf x_0$. This process is computed by
 \begin{equation}\label{eq:ddpm_b}
    \mathbf{x}_{t-1} = \boldsymbol{\mu}_\theta(\mathbf x_t, t) + \sigma_t \mathbf{z}, \quad \mathbf{z} \sim \mathcal{N}(\mathbf{0}, \mathbf{I})
\end{equation}
where $\boldsymbol{\mu}_\theta(\mathbf x_t, t)$ predicts the forward process posterior mean with a network with parameters $\theta$ and $\sigma_t \mathbf{z}$ is its deviation at timestep $t$. We refer the reader to \cite{hoDenoisingDiffusionProbabilistic2020} for an in-depth explanation.

Denoising Diffusion Implicit Models (DDIMs) \cite{songDenoisingDiffusionImplicit2020} enhance DDPMs by introducing a deterministic approach, while allowing optimization towards the same objective. It modifies the reverse process to only depend on the input and the noise prediction network $\epsilon_\theta(x_t, t)$, directly computing the denoised observation $\nnDae$:
\begin{equation}\label{eq:ddim_denoised_obs}
\nnDae(\mathbf{x}_t, t) = \frac{1}{\sqrt{\alpha_t}} (\mathbf{x}_t - \sqrt{1 - \alpha_t} \mathbf{\epsilon}_{\theta}(\mathbf{x}_t, t))
\end{equation}
 and resulting in the inference distribution
\begin{equation}\label{eq:ddim_infer}
   q(\mathbf{x}_{t-1} | \mathbf{x}_t, \nnDae(\mathbf x_t, t)) = \mathcal{N} \left( \sqrt{\alpha_{t-1}} \nnDae(\mathbf x_t, t) + \sqrt{1-\alpha_{t-1}} \frac{\mathbf{x}_t - \sqrt{\alpha_t} \nnDae(\mathbf x_t, t)}{\sqrt{1-\alpha_t}}, \mathbf{I} \right)
\end{equation}.

% \begin{equation}\label{eq:ddim_rev}
% \mathbf{x}_{t-1} = \sqrt{\alpha_{t-1}} \left( \frac{\mathbf{x}_t - \sqrt{1-\alpha_t} \boldsymbol{\epsilon}_\theta(\mathbf{x}_t, t)}{\sqrt{\alpha_t}} \right) + \sqrt{1-\alpha_{t-1}} \boldsymbol{\epsilon}_\theta(\mathbf{x}_t, t)
% \end{equation}

Because it is deterministic, the noise map $\mathbf x_T$ can be used to reconstruct images back to their original state, which implies inversion capabilities. This property is suitable for image editing tasks \cite{huberman-spiegelglasEditFriendlyDDPM2024} and is advantageous for the generation of accurate counterfactuals. 

\subsubsection{Diffusion Autoencoders (Diff-AE).}
A Diffusion Autoencoder (Diff-AE) \cite{preechakulDiffusionAutoencodersMeaningful2022} is an extension of the DDIM and learns a meaningful and decodable representation of an image while leveraging diffusion models for high-quality reconstruction. It encodes an image into a lower-dimensional \emph{semantic subcode} $\zsem$ that captures high-level structured information and a \emph{stochastic subcode} $\mathbf x_T$ that encodes fine-grained stochastic details, taking advantage of the deterministic generative process of the DDIM. The DDIM serves as both the encoder for the stochastic code and the decoder for image reconstruction.

The semantic latent code $\zsem$ is computed by the learnable encoder function $\text{Enc}_{\phi}$ with parameters $\phi$ 
\begin{equation}
    \zsem = \enc(\mathbf x_0)
\end{equation}
mapping an input image $\mathbf x_0$ to its high-level semantic representation. The reverse process of the DDIM is then conditioned on this semantic latent code. The authors extend the expression of the denoised observation $\nnDae$ in Equation \ref{eq:ddim_denoised_obs} and the inference distribution in Equation \ref{eq:ddim_infer} to:
\begin{equation}
\nnDae(\mathbf{x}_t, t, \zsem) = \frac{1}{\sqrt{\alpha_t}} (\mathbf{x}_t - \sqrt{1 - \alpha_t} \mathbf{\epsilon}_{\theta}(\mathbf{x}_t, t, \zsem))
\end{equation}.

To compute the complete denoising process, we use the mean of the inference distribution $q(\mathbf{x}_{t-1} | \mathbf{x}_t, \nnDae(\mathbf x_t, t, \zsem))$ for $1 \leq t < T$ and refer to the complete reconstruction at $t=1$ as
 \begin{equation}
   \hat{\mathbf{x}} = \nnDae(\mathbf{x}_1, 1, \zsem)
\end{equation}.

% It models the distribution $p_\theta(\mathbf x_{t-1} | \mathbf x_t , \zsem)$ to align with the reverse generative process $q(\mathbf x_{t-1}|\mathbf x_t, \mathbf x_0)$ resulting in:

% \begin{align}
% p_{\theta}(\mathbf{x}_{0:T} | \zsem) &= p(\mathbf{x}_T) \prod_{t=1}^{T} p_{\theta}(\mathbf{x}_{t-1} | \mathbf{x}_t, \zsem) \\
% p_{\theta}(\mathbf{x}_{t-1} | \mathbf{x}_t, \zsem) &= 
% \begin{cases} 
% \mathcal{N}(\mathbf{f}_{\theta}(\mathbf{x}_1, 1, \zsem), 0) & \text{if } t=1 \\ 
% q(\mathbf{x}_{t-1} | \mathbf{x}_t, \mathbf{f}_{\theta}(\mathbf{x}_t, t, \zsem)) & \text{otherwise} 
% \end{cases}
% \end{align}

% where the function $\nnDae$ ... \todo{does what exactly? One step of the reverse process perhaps?}

% If $t=1$ this will result in a completely denoised image, effectively reconstructing an input $\mathbf x$:
% \begin{equation}
%    \hat{\mathbf{x}} = \nnDae(\mathbf{x}_1, 1, \zsem)
% \end{equation}

% \begin{equation}
% \mathbf{x}_{t+1} = \sqrt{\alpha_{t+1}} \mathbf{f}_{\theta}(\mathbf{x}_t, t, 
% \zsem) + \sqrt{1 - \alpha_{t+1}} \mathbf{\epsilon}_{\theta}(\mathbf{x}_t, t, \zsem)
% \end{equation}

The stochastic code $\mathbf x_T$ is obtained by rewriting the generative process such that it predicts the last noise state \cite{preechakulDiffusionAutoencodersMeaningful2022}:
\begin{equation}
\mathbf{x}_{T} = \sqrt{\alpha_{T}} \mathbf{f}_{\theta}(\mathbf{x}_{T-1}, {T-1}, 
\zsem) + \sqrt{1 - \alpha_{T}} \mathbf{\epsilon}_{\theta}(\mathbf{x}_{T-1}, T-1, \zsem)
\end{equation}
As $\zsem$ is a lower-dimensional vector, it posesses a limited capacity to encode stochastic details. Therefore, $\mathbf x_T$ is inclined to represent information that was not covered by $\zsem$, focusing on encoding local variations to optimize the training objective. 

Preechakul \etal \cite{preechakulDiffusionAutoencodersMeaningful2022} use the semantic latent code $\zsem$ to allow for semantic editing of images. Even when facial features are fit using only linear models, alterations in $\zsem$ produce meaningful transitions. This property forms the basis for generating counterfactuals on this latent code. 

% \todo{mention if the image is outside of the data manifold, then while the reconstruction is ok any movement in the latent space will result in horrendous images}

\subsection{Adversarial Counterfactual Explanations (ACE)}

Jeanneret \etal \cite{jeanneretAdversarialCounterfactualVisual2023} utilize the DDPM as a regularizer to transform adversarial attacks into semantically meaningful counterfactual explanations. Adversarial Counterfactual Explanations (ACE) generate counterfactual images by optimizing adversarial perturbations in the image space while filtering high-frequency and out-of-distribution artifacts using a diffusion model. 

More specifically, consider $\lossClass(\mathbf x, y)$ as a function that quantifies the match between a sample $\mathbf x$ and a class $y$, typically the cross-entropy loss, which we aim to minimize.
Consider a \emph{filtering function} $F$ that constrains a counterfactual $\mathbf x'$ to the data manifold of the training images. ACE implements this filtering function using the DDPM. It forward and reverse processes the counterfactual (Eq. \ref{eq:ddpm_f} and \ref{eq:ddpm_b}) only to a certain depth $t = \tau$ with $1 < \tau < T$.
This function is then referred to as $F_\tau$. Applied to the optimization problem, we arrive at $\lossClass(F_\tau(\mathbf x'), y)$. The adversarial attack is applied through $F_\tau$, filtering unwanted artifacts, and keeping the structure of the image in tact. The result of the optimization is called the \textit{pre-explanation}. 

To further refine the pre-explanation, the authors use the RePaint in-painting strategy  \cite{lugmayrRePaintInpaintingUsing2022}. Pixel changes are first thresholded with a binary mask to only include the areas highest in magnitude. Then the transitions between the original image and the counterfactual are smoothed out using the DDPM.

ACE provides coherent explanations without requiring modifications to the examined classifier. It surpasses state-of-the-art counterfactual generation techniques across multiple datasets in terms of validity, sparsity, and realism.

\subsection{Diffusion Counterfactuals for Image Regressors}
\begin{figure}
  \centering
  \includegraphics[width=\columnwidth]{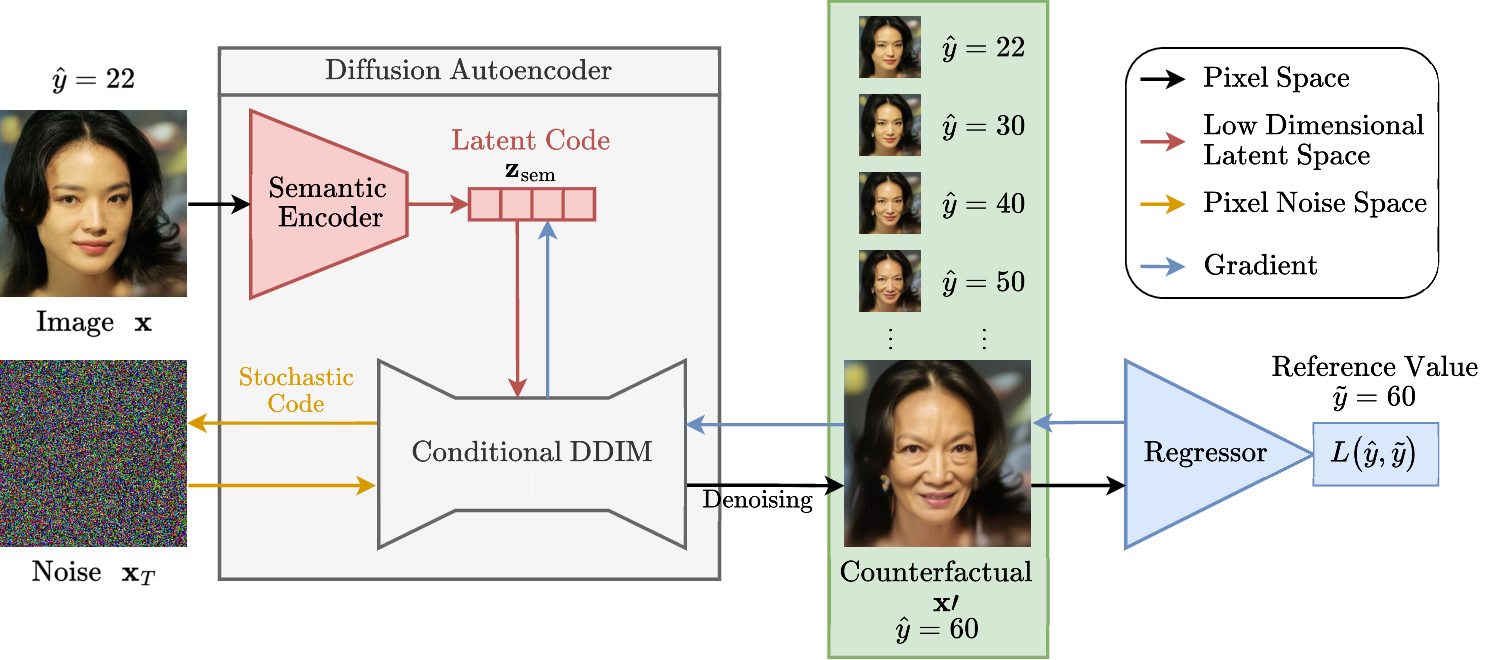}
  \caption{
        \textbf{Diff-AE Regression Explanations (Diff-AE-RE).} The Diff-AE encodes an image into two distinct vectors: high-level semantics in the latent code $\zsem$ and low-level features in the stochastic code $\mathbf x_T$. The adversarial attack creates the gradients that only flow towards the latent code. Diff-AE-RE allows for a smooth transition between the counterfactuals across regression values offering fine-grained inspection of changed features
  }
  \label{fig:dae}
\end{figure}

We present the key contribution of this paper: The generation of counterfactual images tied to a regression problem using diffusion models. In Section \ref{sec:related_work} we discuss two CE generation methods using diffusion-based models: pixel-based and latent space-based. For the pixel space problem we adapt ACE \cite{jeanneretAdversarialCounterfactualVisual2023}, and for the semantic space we use the Diff-AE \cite{preechakulDiffusionAutoencodersMeaningful2022}. Figure \ref{fig:dae} presents an overview of the Diff-AE-based method. We refer the reader to \cite{jeanneretAdversarialCounterfactualVisual2023} for an overview of ACE.

Both methods introduce a loss function targeting the reference value $\yref$ as described in Section \ref{sec:related_work}, along with a measure of difference from the reference value. We use the Mean Square Error (MSE) for its commonality in regression tasks and the Mean Absolute Error (MAE) for its clarity in interpretation. 

Let $\nnReg$ be a regression model with parameters $\psi$, $\mathbf x$ be a data point, $\mathbf x'$ its current counterfactual, initialized to be a copy of $\mathbf x$. The initial prediction of the regressor is given by $\ypred = \nnReg(\mathbf x')$ and we assign a designated reference value $\yref$. The regression loss function $\lossReg(\ypred, \yref) = \operatorname{MSE}(\ypred, \yref)$ is the objective of adversarial attacks and $\confidence(\ypred, \yref) = \operatorname{MAE}(\ypred, \yref)$ is the difference to the reference value. We empirically choose an early stopping criterion $\confidence \leq 0.05$, similar to the validation error of CelebA-HQ dataset regressors. It balances explanation proximity to $\yref$, while limiting optimization steps.

Hence, to produce the counterfactuals $\mathbf x'$ we perform the adversarial attack and optimize
\begin{equation}
\argmin_{\mathbf x'} \lossReg(\ypred, \yref) \quad\text{s.t.}\quad \confidence(\ypred, \yref) \leq 0.05
\end{equation}

To assess the performance of our methods, we use a novel, regression specific metric and established metrics used in ACE \cite{jeanneretAdversarialCounterfactualVisual2023} and STEEX \cite{jacobSTEEXSteeringCounterfactual2022}. We elaborate on the specific details of these metrics in Section \ref{sec:experiments}.

\subsubsection{Adversarial Counterfactual Regression Explanations.} 
We adapt ACE using the regression loss described previously $\lossReg$. This algorithm performs the prediction on the pre-explanation, $\ypred = \nnReg(F_\tau(\mathbf x'))$. The optimized loss function becomes $\lossReg(\ypred, \yref)$, using the mentioned early stopping criterion. The final refinement stage is performed after a successful adversarial attack. We refer to this regression-specific adaption as \emph{AC-RE}. We expect the same properties of ACE to apply to AC-RE as well, namely semantic changes in the images on the data manifold with filtered out-of-distribution noise. In contrast to ACE, we cannot use a distance function for enhanced sparsity. During empirical testing, we found that the distance function is too limiting to produce counterfactuals that change the predicted value. 

\subsubsection{Diff-AE Regression Explanations.} To adapt Diff-AE we use only the semantic code $\zsem$ to optimize toward the reference value. For the counterfactual $\mathbf x'$ we produce its reconstruction $\hat{\mathbf x}'$ using the denoising process explained in Section \ref{sec:diffusion}. With this, the predicted regression value becomes $\ypred = \nnReg(\hat{\mathbf x}')$. We then perform an adversarial attack w.r.t. $\zsem$ using the regression loss $\lossReg(\ypred, \yref)$ until we reach the early stopping criterion. We call this method \emph{Diff-AE-RE}. As we are only attacking the latent space of $\zsem$, we aim to change only the high-level features of the images. 

To enforce semantic sparsity, we add a distance function $d$ between the latent code of the input image and the counterfactual to the objective of the adversarial attack. Through empirical tests elaborated in Section \ref{sec:ablation}, we choose the distance function to be the $\ell_1$-distance with its regularization constant $\lambda_d = 10^{-5}$. Let $\zsem$ be the latent code of the input image $\mathbf x$ and $\mathbf z'_\text{sem}$ the latent code of the counterfactual $\mathbf x'$. We arrive at the following extended objective:

\begin{equation}
\argmin_{\mathbf \zsem'} \lossReg(\ypred, \yref) + \lambda_d\ d(\zsem, \mathbf z'_\text{sem}) \quad\text{s.t.}\quad \confidence(\ypred, \yref) \leq 0.05
\end{equation}

This approach is similar to and heavily inspired by approximate diffeomorphic counterfactuals \cite{dombrowskiDiffeomorphicCounterfactualsGenerative2024}, which demonstrate strong theoretical properties for autoencoders. We suggest an analysis of the implications for Diff-AE as future work.

% 4. Experiments
\section{Experiments}\label{sec:experiments}
We experimentally evaluate the effectiveness of our methods in both synthetic and real-world scenarios. We compare their performance against classifier-based approaches and analyze their ability to provide meaningful regression-specific explanations.

First, we describe the datasets used in our experiments, which include a synthetic dataset and CelebA-HQ. We elaborate on the metrics evaluated on these datasets, covering validity, realism, and sparsity.
We then outline the implementation details of our methods, specifying how we use the diffusion and regression models.
Next, we present quantitative results, comparing our algorithms with classifier-based counterfactual generation approaches. We complement this with qualitative results, providing visual examples of our generated counterfactuals. Additional examples illustrate our models' ability to reveal spurious correlations and produce fine-grained regression-specific explanations. 
Finally, we conduct an ablation study that examines the contributions of key components to our methods.

\subsection{Datasets}

We first test the algorithms on a synthetic dataset called the square dataset. It consists of $64\ \times\ 64$ images that have a uniform background from black to white and contains a square of size $8\ \times\ 8$ in varying shades of red, enclosed by a gray border. The regression task is to predict the red intensity of the square. This results in a latent space with two dimensions that we can directly control. A minor amount of random Gaussian noise is added to the images to serve as regularization. The objective is to achieve a gradual transition in the color of the square, either from dark to bright red or the reverse. We choose the reference value inversely to the square's red intensity: darker red squares have bright red reference values, and vice versa. This ensures a balanced contrast for evaluating transitions. By construction, this dataset allows for precise control of particular evaluation aspects, as detailed in Section \ref{sec:eval_methods}.

Following recent literature on generative CE \cite{jacobSTEEXSteeringCounterfactual2022,jeanneretDiffusionModelsCounterfactual2022, jeanneretAdversarialCounterfactualVisual2023}, we evaluate our algorithms using the CelebA-HQ dataset \cite{karrasProgressiveGrowingGANs2018}. It contains images of size $256\ \times \ 256$ of cropped human faces. We choose the attribute ``age'' to generate the counterfactuals. If a face is marked as ``young'' we choose the reference value $\yref = 80$, while we assign "old" $\yref = 10$ to create CEs that are sufficiently distinct. We normalize ages to a 0-1 range for model training and evaluation.

To train a regression model for age prediction, we use the imdb-wiki-clean dataset \cite{lin2021fpage}, a cleaned version of the imdb-wiki dataset \cite{rotheDEXDeepEXpectation2015}. It consists of more than 500,000 face images with gender and age labels, with the latter being the target of the regressor. The images were cropped to match the CelebA-HQ format.

\subsection{Evaluation Methods}\label{sec:eval_methods}

We follow Jeanneret \etal \cite{jeanneretAdversarialCounterfactualVisual2023} and assess CEs validity, sparsity, and realism.

The \textit{validity of the explanations} for classification tasks is commonly measured by the flip rate of the CE produced. As there are no fixed thresholds for regression tasks, we choose the oracle score \cite{hvilshojQuantitativeEvaluationsCounterfactuals2021} and adapt it to the regression case. We introduce a secondary regressor with the same architecture as an oracle and compute the MAE between the predictions of the original regressor and the oracle. We refer to this as the \textit{Oracle MAE}, suggesting the adversarial potency between the chosen models. A low score indicates the semantic and model-independent nature of the generated counterfactuals.

We determine \textit{sparsity or proximity} by measuring changes in the 40 binary attributes of the CelebA-HQ dataset between the input image and the CE. The VGGFace2 model \cite{caoVGGFace2DatasetRecognising2018} predicts all attributes of the image pair and computes the mean number of attributes changed (MNAC). Like Jeanneret \etal \cite{jeanneretAdversarialCounterfactualVisual2023}, we use face verification accuracy (FVA) \cite{caoVGGFace2DatasetRecognising2018} and Face Similarity (FS) \cite{jeanneretAdversarialCounterfactualVisual2023} to assess facial similarity through the deep features of the VGGFace2 model \cite{caoVGGFace2DatasetRecognising2018}. FS is defined as the cosine distance between the deep features of a pair of images, while FVA thresholds FS to create a binary output.

We evaluate the \textit{realism} of the counterfactual images employing the commonly used FID \cite{heuselGANsTrainedTwo2017} between the set of input images and the respective CEs. %\todo{any comments regarding FID score and how it might be biased?}

For the square dataset, we evaluate CE with the same categories of metrics. However, since we can directly control the underlying parameters of the dataset, we can directly extract the MAE of the color values of the square and the background. The first evaluates validity, while the second evaluates sparsity. To assess realism, we also use the FID.

\subsection{Implementation Details}

This subsection covers technical details of our two approaches and the utilized regression models. We make our code and models available on \href{https://github.com/DevinTDHa/Diffusion-Counterfactuals-for-Image-Regressors}{GitHub}.

\subsubsection{Algorithms.} To reduce the computational requirements for propagating gradients through the iterative process of the DDPM, ACE employs a "time-step re-spacing mechanism" \cite{jeanneretAdversarialCounterfactualVisual2023}. This approach reduces the number of steps needed at the cost of decreased quality. Hence, AC-RE uses the same hyperparameters as ACE, using 25 steps to noise and 5 steps to de-noise the image, i.e. $\tau=5$. We reuse the CelebA-HQ DDPM by ACE for AC-RE. To optimize the counterfactual objective, we follow ACE and use projected gradient descent \cite{madryDeepLearningModels2019} with a learning rate of $\frac{1}{255}$.

Similarly, we balance this tradeoff with Diff-AE-RE. By default, the underlying DDIM uses 250 implicit forward steps and 20 backward steps. Through empirical observations, we choose $T=10$ for the reverse process to maintain high quality while allowing larger batch sizes. We use the pre-trained Diff-AE by Preechakul \etal \cite{preechakulDiffusionAutoencodersMeaningful2022}, which was trained on FFHQ \cite{Karras_2019_CVPR}. The cropping methods of FFHQ and CelebA-HQ overlap, allowing us to use the model without further modifications. We optimize the counterfactual objective using Adam \cite{kingmaAdamMethodStochastic2017} with a learning rate of $0.002$, further elaborated in Section \ref{sec:ablation}.

\subsubsection{Regression Models.} To train the regression models, we fine-tune the pre-trained classification model used by ACE \cite{jeanneretAdversarialCounterfactualVisual2023}, which is based on DenseNet \cite{huangDenselyConnectedConvolutional2017}. To adapt this model to a regression task, we replace the final classification layer of the network with a fully connected layer with a single continuous output while freezing all other model weights during training. Hence, the model reuses its feature extraction capabilities, creating explanations comparable to those of the related works.

We also base the oracle model for the Oracle MAE on the DenseNet architecture. Hvilshoj \etal \cite{hvilshojQuantitativeEvaluationsCounterfactuals2021} argue that the choice of oracle has a major influence on the final score. We aim to ensure comparability with related works by selecting the same architecture for the oracle. However, adversarial vulnerabilities may transfer between the models if we choose to freeze layers. Hence, we decide to fine-tune the entire model, minimizing this risk.

\subsection{Quantitative Comparisons} \label{sec:quant_res}

% ---------------- SQUARE ----------------
We analyze our methods with the metrics described in Section \ref{sec:eval_methods}. First, we examine the results of the synthetic square data. Afterward, we assess how the methods compare to the classifier-based methods on CelebA-HQ.

Table \ref{tab:quant_square} shows the results for the square dataset. We observe that AC-RE generally performs better in this task. It produces much sparser CE due to the limitations that ACE imposes on the changes: The structure of the image is mostly intact because of the lower amount of noising, and the final refinement stage further limits the areas that are changed.

In contrast, Diff-AE-RE reconstructs the image only from the semantic and stochastic code from a completely noised image. Because of this, the CEs lose their noise pattern, resulting in a higher FID. Figure \ref{fig:qual_res} shows such an example. Moreover, as the structure of the image is expressed with the semantic code, alterations to it naturally lead to broader changes than AC-RE's pixel-based approach.

\begin{table}
\caption{\textbf{Quantitative Results for the Square dataset.} We show the best performance in bold. AC-RE performs much better in sparsity than Diff-AE-RE} \label{tab:quant_square}
\centering
\begin{tabular}{l|ccc}
\hline
Method & Square MAE & FID & Background MAE  \\
\hline
AC-RE & \textbf{0.13} & \textbf{38.1} & \textbf{0.012} \\
Diff-AE-RE & 0.167 & 121.9 & 0.032 \\
\hline
\end{tabular}
\end{table}

% ---------------- CelebaHQ ----------------
Table \ref{tab:quant_res} shows the CelebA-HQ results for the attribute ``age'' with the evaluation metrics discussed in Section \ref{sec:eval_methods} and compares them to related classification methods.
Compared to the classification-based methods, the FID of our methods is much higher. This is expected; since the CE for classification only need to cross the decision boundary to be considered successful, regression CE need to traverse a significant distance on the loss surface to approach the chosen reference values. Despite this, AC-RE still produces sparse CEs, indicated by the low MNAC score and the high facial similarity indicated by FVA and FS.

In contrast, Diff-AE-RE seems to produce CEs that are more semantically meaningful and realistic, indicated by the lower Oracle MAE and FID scores. Diff-AE-RE is limited to modifications in the latent space, restricting non-semantic adversarial components in the explanations and enabling wider changes in facial structure and properties. 

This factor significantly decreases the performance of FVA and FS in the specific task of age regression. In Figure \ref{fig:qual_res}, we present CEs from Diff-AE-RE that provide more convincing explanations than AC-RE when creating a CE for an old person turning young. Aging significantly alters facial structure and skin color due to numerous biological processes. Representing these age-related changes in the counterfactual results in larger pixel changes. This affects the extracted deep facial features of the VGGFace2 model, thus lowering the scores.

\begin{table}
\centering
\caption{\textbf{Quantitative Results for CelebA-HQ.} We show the best-performing regression-based method in bold. For reference, we compare the quantitative results of ACE \cite{jeanneretAdversarialCounterfactualVisual2023} on the ``age'' attribute and show them in italics. Our methods perform similarly to classifier-based methods. AC-RE again produces much sparser CEs, keeping facial features intact, while Diff-AE-RE produces CEs that are more semantically sound and realistic} 
\begin{tabular}{l|ccccc}
\hline
Method & Oracle MAE & FID & FVA & FS & MNAC \\
\hline
DiVE \cite{rodriguezTrivialCounterfactualExplanations2021} & - & \textit{33.8} & \textit{98.2} & - & \textit{4.58} \\
DiVE$^{100}$
\cite{rodriguezTrivialCounterfactualExplanations2021} & - & \textit{39.9} & \textit{52.2} & - & \textit{4.27} \\
STEEX \cite{jacobSTEEXSteeringCounterfactual2022} & - & \textit{11.8} & \textit{97.5} & - & \textit{3.44} \\
DiME \cite{jeanneretDiffusionModelsCounterfactual2022} & - & \textit{4.15} & \textit{95.3} & \textit{0.6714} & \textit{3.13} \\
ACE $\ell_1$ \cite{jeanneretAdversarialCounterfactualVisual2023} & - & \textit{1.45} & \textit{99.6} & \textit{0.7817} & 3.20 \\
ACE $\ell_2$ \cite{jeanneretAdversarialCounterfactualVisual2023} & - & \textit{2.08} & \textit{99.6} & \textit{0.7971} & 2.94 \\
\hline
AC-RE & 0.260 & 30.9 & \textbf{93.1} & \textbf{0.666} & \textbf{2.68} \\
Diff-AE-RE & \textbf{0.184} & \textbf{26.7} & 80.1 & 0.599 & 3.61 \\
\hline
\end{tabular}
\label{tab:quant_res}
\end{table}

% \begin{itemize}
%     \item Quantitative results shown in Table \ref{tab:quant_res}
%     \item The methods show similar value ranges to their classification counterparts.
%     \item AC-RE seems keep its property of sparseness well.
%     \item Diff-AE-RE has difficulties keeping it sparse (explained due to large changes in the face with age)
% \end{itemize}

\subsection{Qualitative Results} \label{sec:qual_res}
\begin{figure}
    \centering
    \newcommand{\rotated}[1]{\rotatebox[origin=c]{90}{#1}}
    \newcommand{\rotatedRaised}[2]{\raisebox{#1}{\rotatebox[origin=c]{90}{#2}}}
    \newcommand{\includeadjustedgraphic}[1]{\adjustbox{valign=m}{\includegraphics[width=\linewidth]{#1}}}
    % Define a new column type for fixed 5% width columns (first two and last column)
    \newcolumntype{F}{>{\centering\arraybackslash}m{\dimexpr0.05\linewidth\relax}}
    % Define a new column type for fixed 28.33% width columns (columns 3,4,5)
    \newcolumntype{I}{>{\centering\arraybackslash}m{\dimexpr0.3\linewidth\relax}}
    \begin{tabular}{FIIIF}
        
        % Header
        & Square & CelebA-HQ & CelebA-HQ \\
        &
        $\yref = 0$ \hspace{0.9cm} $\yref=1$ &
        Young to Old: $\yref = 80$ &
        Old to Young: $\yref = 10$ \\

        % Original Image
        \rotated{$\mathbf x$} &
        \includegraphics[width=\linewidth]{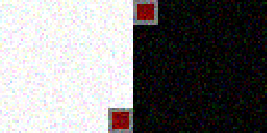} &
        \includegraphics[width=\linewidth]{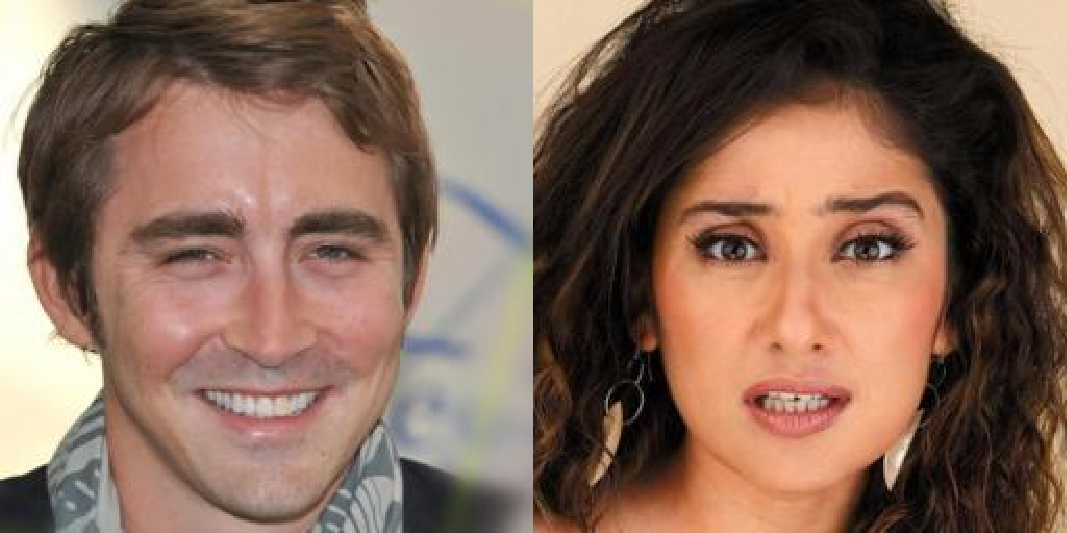} &
        \includegraphics[width=\linewidth]{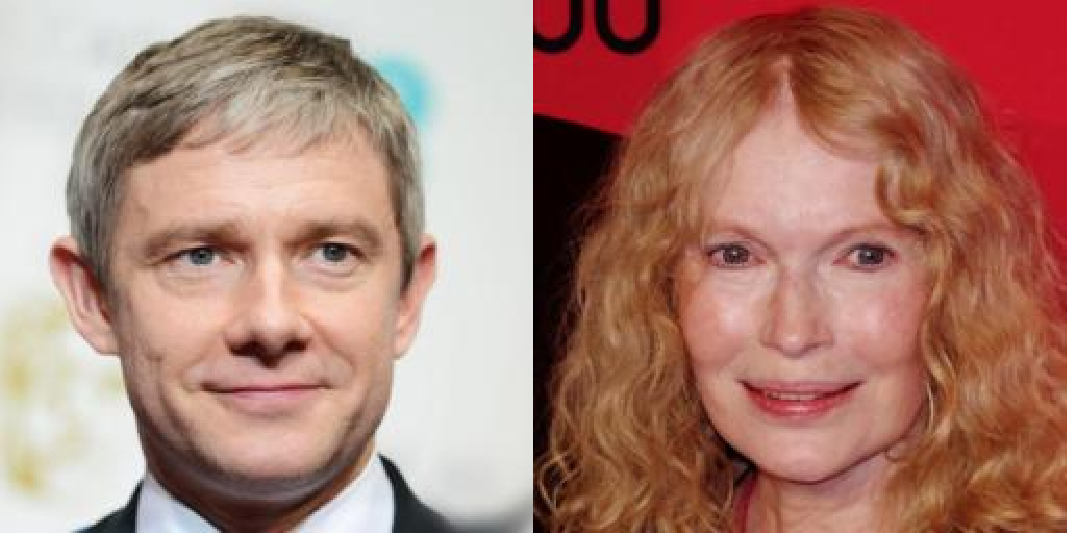} \\

        % Initial Predictions
        &
        $y = 0.49$ \hspace{0.5cm} $y=0.50$ &
        $\ypred = 40$ \hspace{0.7cm} $\ypred=31$ &
        $\ypred = 52$ \hspace{0.7cm} $\ypred=55$ \\
    
        % Ace-R
        \rotatedRaised{0.5cm}{AC-RE} &
        \includegraphics[width=\linewidth]{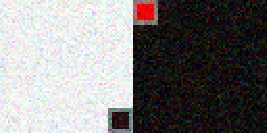} &
        \includegraphics[width=\linewidth]{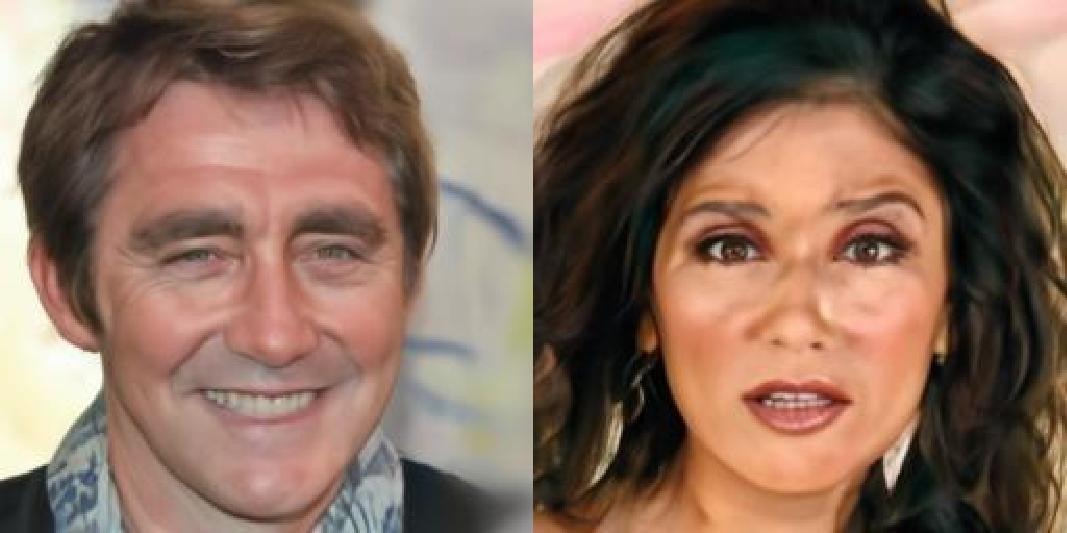} &
        \includegraphics[width=\linewidth]{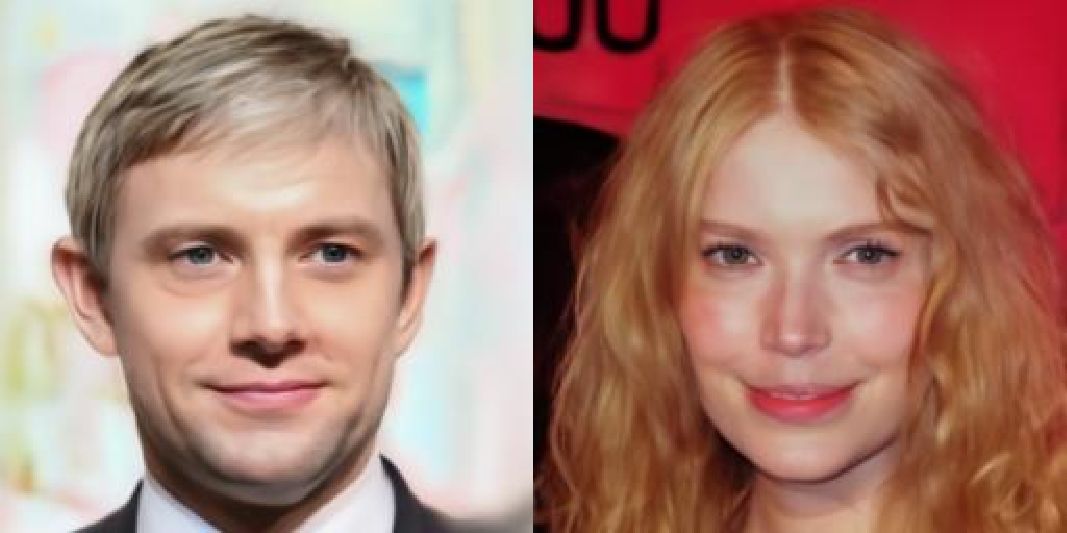} \\

        % AC-RE Predictions
        &
        $y = 0.14$ \hspace{0.5cm} $y=0.92$ &
        $\ypred_o = 60$ \hspace{0.7cm} $\ypred_o=57$ &
        $\ypred_o = 20$ \hspace{0.7cm} $\ypred_o=13$ \\

        % Diff-AE-RE
        \rotatedRaised{0.6cm}{Diff-AE-RE} &
        \includegraphics[width=\linewidth]{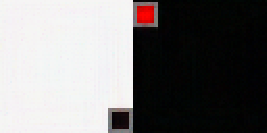} &
        \includegraphics[width=\linewidth]{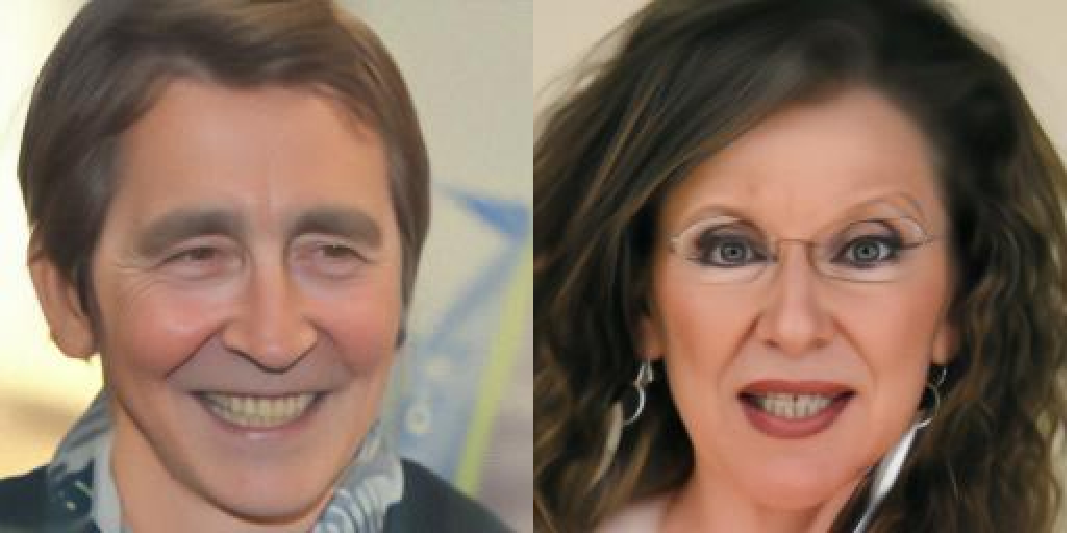} &
        \includegraphics[width=\linewidth]{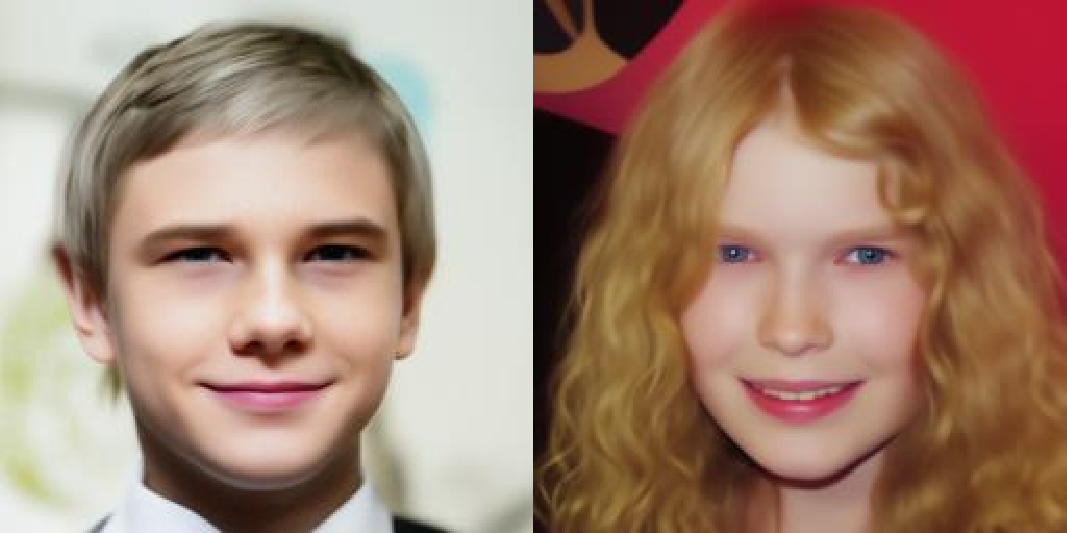} \\
        
        % Diff-AE-RE Predictions
        &
        $y = 0.07$ \hspace{0.5cm} $y=0.85$ &
        $\ypred_o = 69$ \hspace{0.7cm} $\ypred_o=73$ &
        $\ypred_o = 13$ \hspace{0.7cm} $\ypred_o=14$ \\
    
    \end{tabular}

    \caption{ \textbf{Qualitative Results.} The first row shows the input images $\mathbf x$ and their reference values $\yref$, while the following rows show the CEs of AC-RE and Diff-AE-RE. The captions indicate the extracted color of the square $y$, the regressor prediction $\ypred$, or the oracle prediction $\ypred_o$. AC-RE creates sparse explanations, while Diff-AE-RE creates broader and more realistic modifications. For the synthetic dataset, AC-RE creates more accurate counterfactuals. For CelebA-HQ, AC-RE primarily changes the textures in the face. In contrast, Diff-AE-RE alters facial shape, skin and teeth color, and accessories. Its oracle score indicates more realistic changes, aligning closely with the reference values}
    \label{fig:qual_res}
\end{figure}
% - See Figure \ref{fig:qual_res}: comparison of AC-RE and Diff-AE-RE
% - Diff-AE-RE: semantic and real: skin color, face shape, accessories (glasses and earrings), shape of smile and color of the teeth, young: Shape of the eyes
% - AC-RE primarily changes the texture of the image/faces (however is especially strong for the squares)
% - Diff-AE-RE produces changes that are non-sparse in pixel space but perhaps sparse in semantics: general shape of the face, hairline and color, color of the skin or teeth; Addition of glasses
% - transition to spurious correlation part and regression part with glasses and earrings:

We show a selection of qualitative results in Figure \ref{fig:qual_res} for the squares and CelebA-HQ datasets, with their respective reference values. Similarly to the observations of the quantitative results in Section \ref{sec:quant_res}, we see that AC-RE produces sparser CEs, while Diff-AE-RE produces broader changes. 

For the square dataset, AC-RE creates a uniform color for the square and even reconstructs the noise patterns. Diff-AE-RE cannot reproduce these patterns, as previously described in Section \ref{sec:quant_res}. Moreover, the algorithm slightly shifted the location of the square of the image with the darker background (the border is no longer aligned with the edge of the image). Both algorithms perform similarly well regarding altering the actual color of the square.

For CelebA-HQ, AC-RE focuses on texture alterations, such as adding wrinkles and smoothing the skin, staying sparse in the pixel space. Diff-AE-RE introduces more distinct and realistic changes. This is reflected in the predictions of the Oracle model. They align more closely with the reference values, indicating fewer adversarial components. Although these changes are less minimal concerning the changed pixels, we see them as an advantage in creating additional features for interpretability. Specifically, we observe changes in facial shape, skin, and teeth color, and the presence of accessories.

The final point stands out significantly in our observations. For many CEs, we observe that Diff-AE-RE adds glasses or earrings. We investigate these phenomena in more detail to uncover spurious correlations in Section \ref{sec:spur} and age-range-specific explanations in Section \ref{sec:regression_specific}.

\subsubsection{Spurious Correlations of the Regressor.} \label{sec:spur}
% \begin{itemize}
%     \item Figure \ref{fig:exp_glasses} shows how we can use the information generated by the CE of the Diff-AE-RE: 
%     \item for lots of images, glasses were added thus we try to influence the decision of the regressor by adding glasses
%     \item done by using an ai image editor to mask areas of an image, then generate via inpainting and a prompt "person wearing glasses"
%     \item experiment was a success: on average 7 years older. With an specially strong example in Figure \ref{fig:exp_glasses}
% \end{itemize}

Counterfactual explanations are the most effective when the changes are minimal and actionable. However, the definition of minimality changes depending on the context \cite{guidottiCounterfactualExplanationsHow2022}. While ACE favors minimal pixel changes \cite{jeanneretAdversarialCounterfactualVisual2023}, enforcing it may not faithfully represent the behavior of the model and limit intuitive counterfactuals \cite{delaneyCounterfactualExplanationsMisclassified2022}.

Although adding an accessory such as glasses introduces a significant pixel footprint in the image, it is a simple feature to understand for humans. As we observed in Section \ref{sec:qual_res}, Diff-AE-RE frequently adds glasses to the faces of CelebA-HQ to create a counterfactual of a young person to appear old. 

\begin{figure}
\centering
    \centering
    \begin{tabular}{cccc} % Two columns, no extra space
    Original $\mathbf x_1$ & Edited $\mathbf x_2$ & Original $\mathbf x_1$ & Edited $\mathbf x_2$\\
    \includegraphics[width=0.24\textwidth]{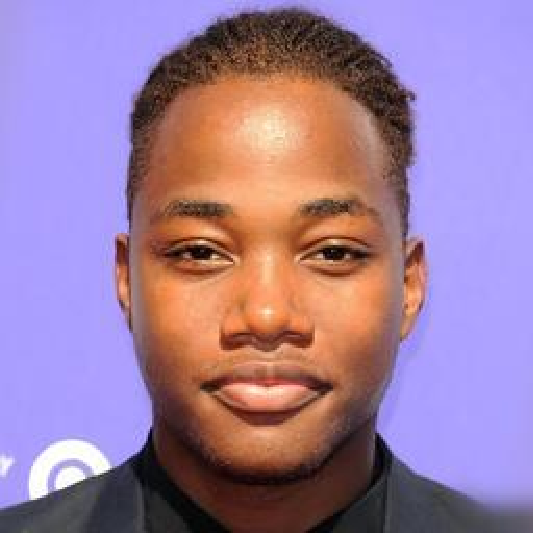} &
    \includegraphics[width=0.24\textwidth]{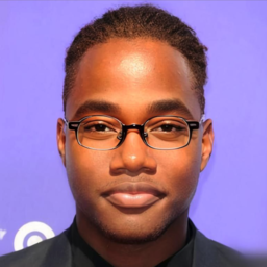} &
    \includegraphics[width=0.24\textwidth]{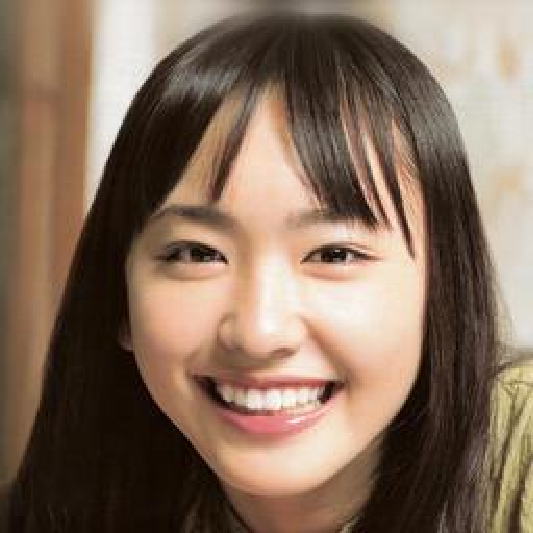} &
    \includegraphics[width=0.24\textwidth]{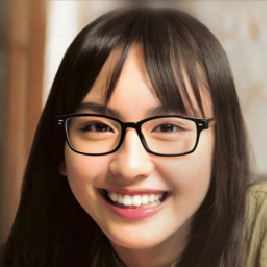} \\
    \multicolumn{2}{c}{\(\hat{y}_{2}-\hat{y}_{1}=10.62\)} & 
    \multicolumn{2}{c}{\(\hat{y}_{2}-\hat{y}_{1}=4.78\)} \\ % Spanning caption
    % second row of images
    \includegraphics[width=0.24\textwidth]{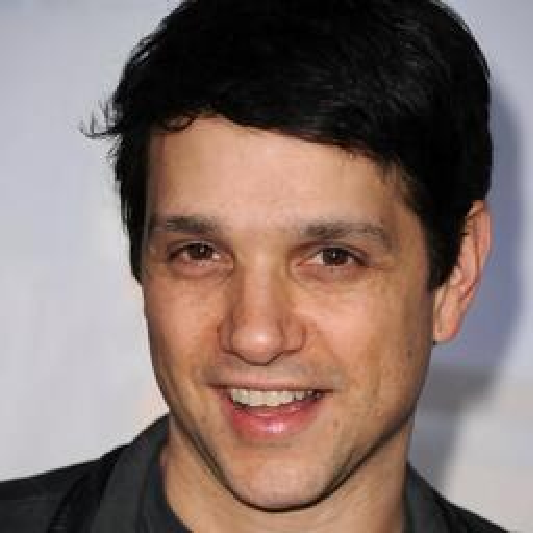} &
    \includegraphics[width=0.24\textwidth]{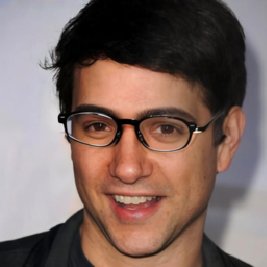} &
    \includegraphics[width=0.24\textwidth]{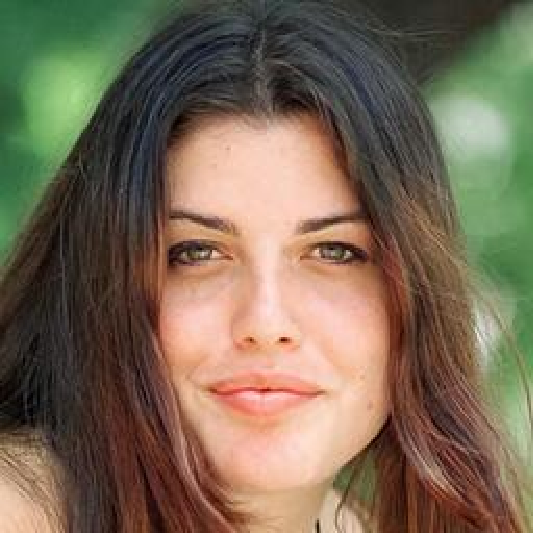} &
    \includegraphics[width=0.24\textwidth]{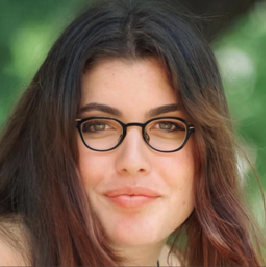} \\
    \multicolumn{2}{c}{\(\hat{y}_{2}-\hat{y}_{1}=6.90\)} & 
    \multicolumn{2}{c}{\(\hat{y}_{2}-\hat{y}_{1}=4.96\)} \\ % Spanning caption
    \end{tabular}

\caption{
    \textbf{Spurious Correlation.} We reveal a spurious correlation in the regressor for sample images from the CelebA-HQ validation set. We use the getimg.ai Image Editor \cite{ImageEditorGetimgai} to add glasses to each person via inpainting. On average, the person appears to be 7 years older than the model, with the first example showing an especially strong change. The effect seems to be more pronounced for male faces
}
\label{fig:exp_glasses}
\end{figure}

Using this information, we conduct an experiment on a subset of the CelebA-HQ validation set: We use the getimg.ai Image Editor \cite{ImageEditorGetimgai} to mask the area around the eyes and inpaint the image with the prompt ``a person wearing glasses''. 
This way, we precisely control the modifications to the image while creating realistic-looking images. Afterward, we compute the difference between the predictions of the regressor for the original and the edited image. We show the results in Figure \ref{fig:exp_glasses}.

We observe that the modifications increase the predicted age by at least about five years, on average seven years. Notably, the edit of the first example has a particularly strong response, increasing the prediction by eleven years. However, the effect seems to be more pronounced for male faces, indicating a potential bias in the training dataset.  Using our counterfactuals, we show that suggested feature changes can be directly applied in a very intuitive way (``putting on glasses'') to alter the result of the regression model.

\subsection{Regression Specific Explanations} \label{sec:regression_specific}
% \begin{itemize}
%     \item Figure \ref{fig:exp_multitarget} shows CE for multiple reference values. Initial prediction is 20 years old.
%     \item granularity of the explanations is an advantage over classifiers: We can see how different value regions impact the regressors decision
%     \item Example interpretation: from 20 to 40 years, the earring and the shape of the mouth / teeth are most important. The older, the person, the more the shape of the face/neck, shape of lips, color of the hair are considered.
% \end{itemize}

To illustrate the differences between classification- and regression-based CEs, we explain a single image for various reference values, showing granular explanations between age regions. We choose five representative reference values $\yref$ in the target domain, specifically 10, 20, 40, 60, and 80, and analyze the main feature changes between each.

\begin{figure}
\centering
\begin{tabular}{@{}m{0.05\textwidth}*6{>{\centering}m{0.14\textwidth}}m{0.05\textwidth}@{}}

% Header
&
$\hat{\mathbf x}, \mathbf x$ &
$\tilde y = 10$ &
$\tilde y = 20$ &
$\tilde y = 40$ &
$\tilde y = 60$ &
$\tilde y = 80$ &
\\

% Row 1
% \multirow{2}{*}{\rotatebox[origin=c]{90}{AC-RE}} &
\multirow{2}{*}{\raisebox{-1cm}{\rotatebox[origin=c]{90}{AC-RE}}} &
% \caption*{$\hat{\mathbf x}, \mathbf x: \hat y = 23$}
\includegraphics[width=\linewidth]{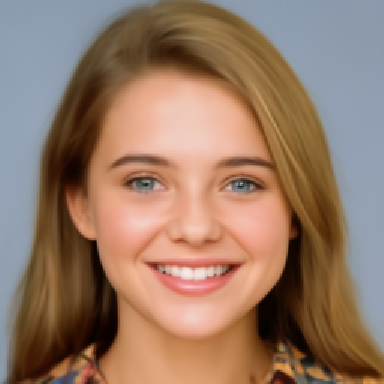} & 
% \caption*{$\tilde y = 10$}
\includegraphics[width=\linewidth]{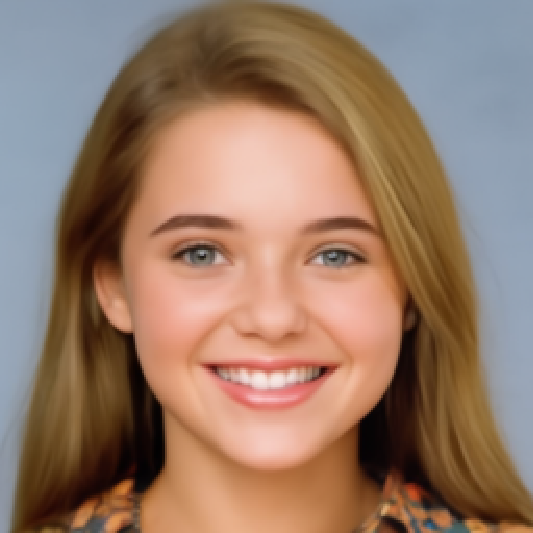} & 
% \caption*{$\tilde y = 20$}
\includegraphics[width=\linewidth]{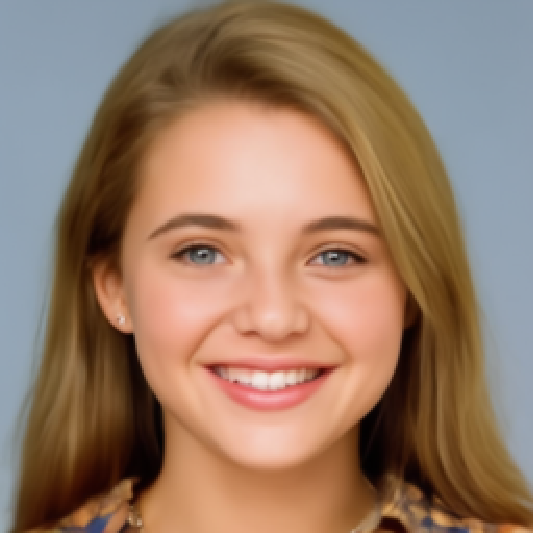} & 
% \caption*{$\tilde y = 40$}
\includegraphics[width=\linewidth]{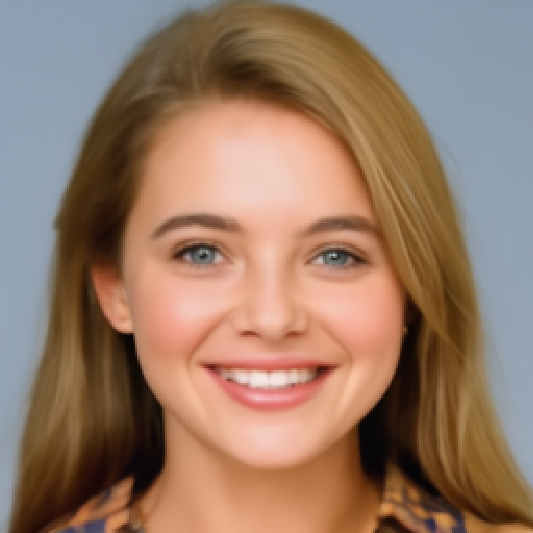} & 
% \caption*{$\tilde y = 60$}
\includegraphics[width=\linewidth]{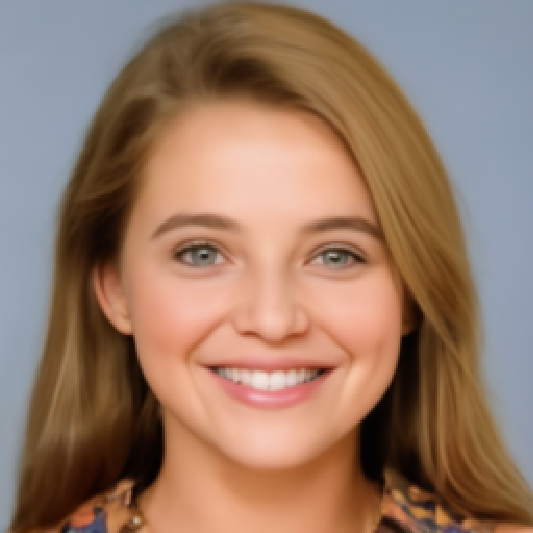} & 
% \caption*{$\tilde y = 80$}
\includegraphics[width=\linewidth]{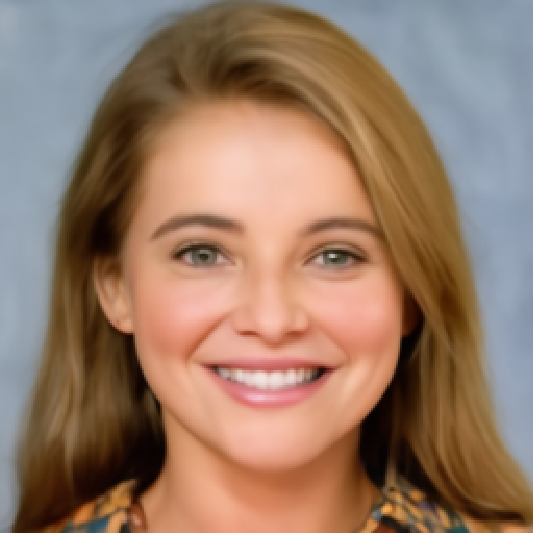} & 
\\ % Empty cell

% Row 2
% Desc from previous  
& 
\includegraphics[width=\linewidth]{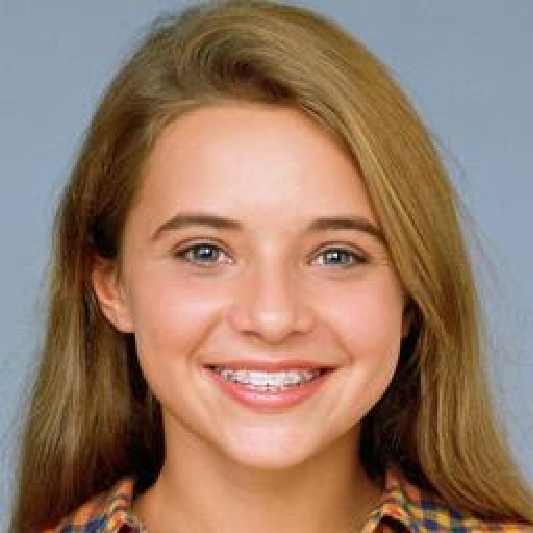} & 
\includegraphics[width=\linewidth]{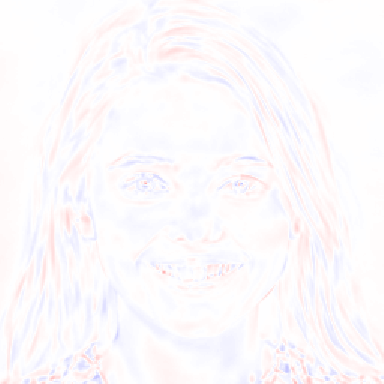} & 
\includegraphics[width=\linewidth]{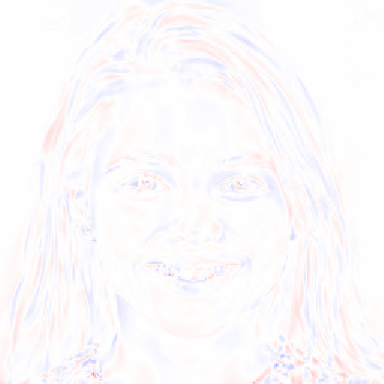} & 
\includegraphics[width=\linewidth]{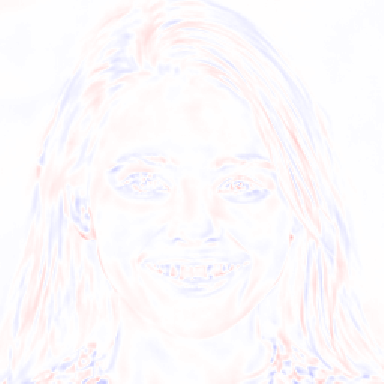} & 
\includegraphics[width=\linewidth]{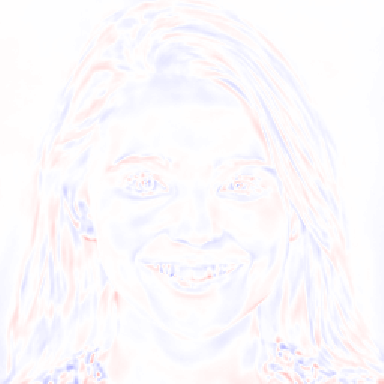} & 
\includegraphics[width=\linewidth]{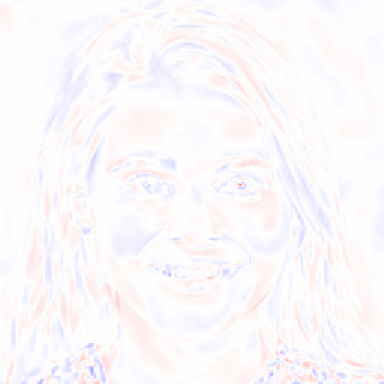} & 
\includegraphics[height=1.7cm]{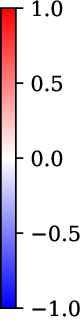} \\

% Row 3
\multirow{2}{*}{\raisebox{-\height}{\rotatebox[origin=c]{90}{Diff-AE-RE}}} &
\includegraphics[width=\linewidth]{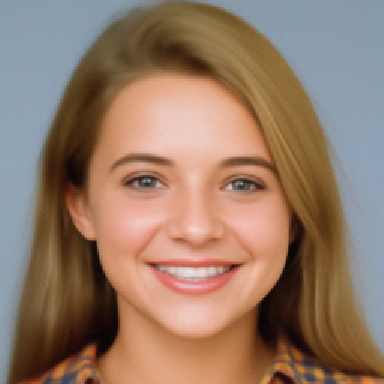} & 
\includegraphics[width=\linewidth]{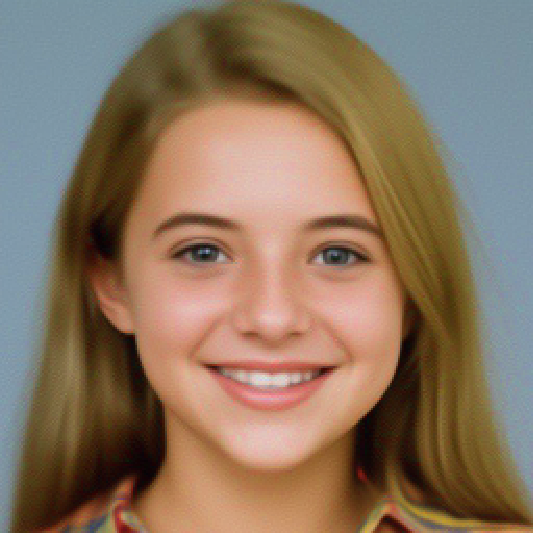} & 
\includegraphics[width=\linewidth]{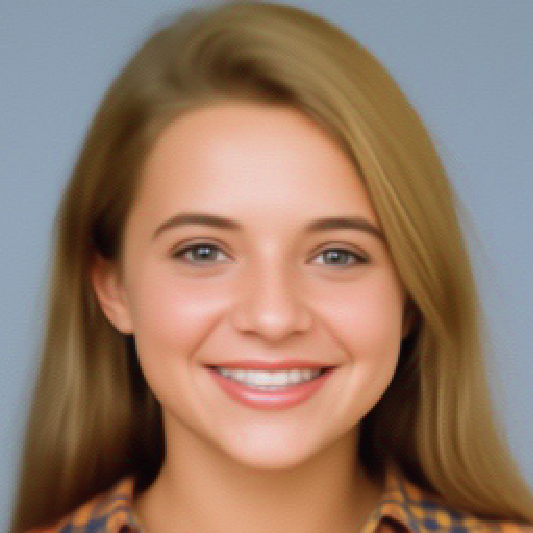} & 
\includegraphics[width=\linewidth]{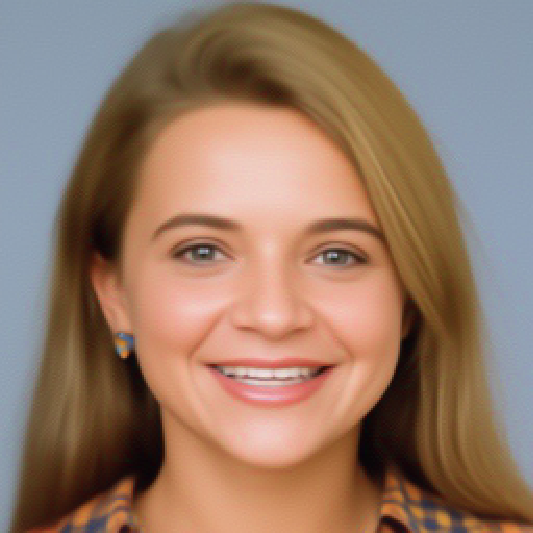} & 
\includegraphics[width=\linewidth]{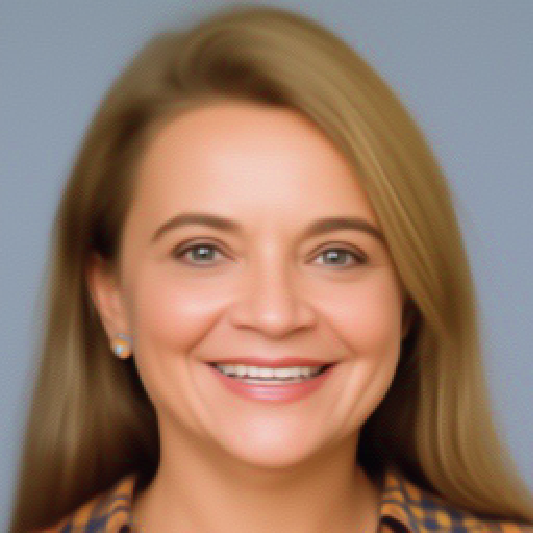} & 
\includegraphics[width=\linewidth]{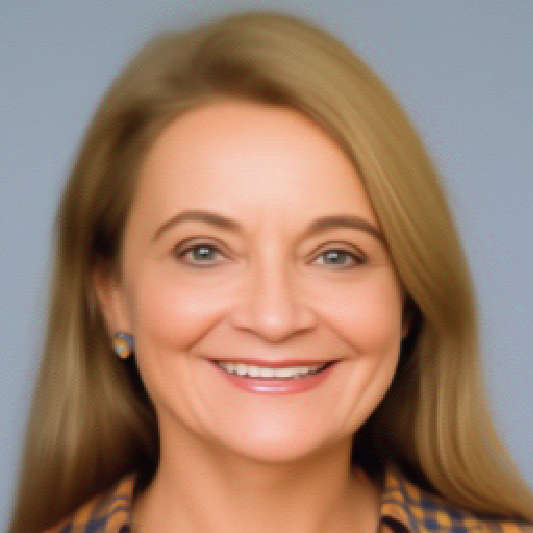} & 
\\ % Empty cell

% Row 4
% Desc from previous  
& 
\includegraphics[width=\linewidth]{figures/qual_res/multitarget/comparison/nm0003115_rm2138097664_1979-1-25_1991.jpg.eps} & 
\includegraphics[width=\linewidth]{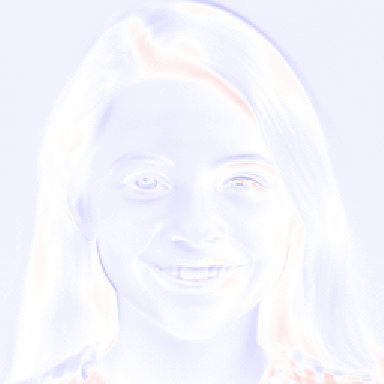} & 
\includegraphics[width=\linewidth]{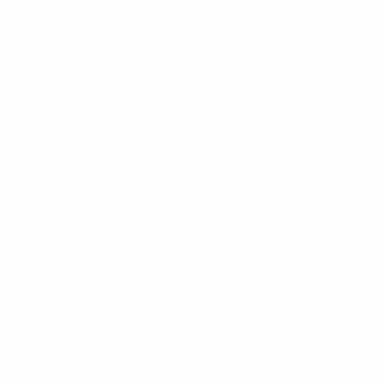} & 
\includegraphics[width=\linewidth]{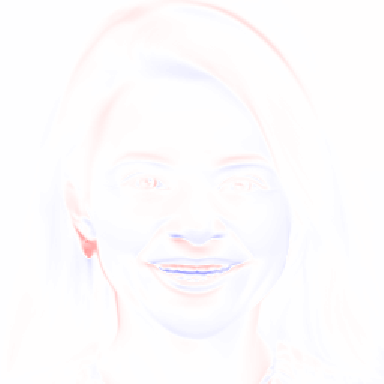} & 
\includegraphics[width=\linewidth]{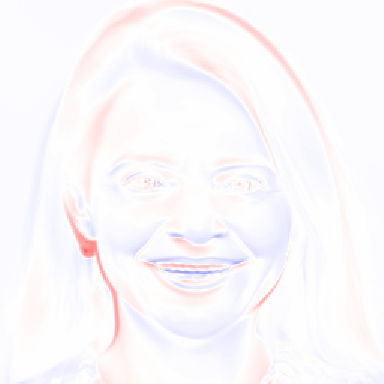} & 
\includegraphics[width=\linewidth]{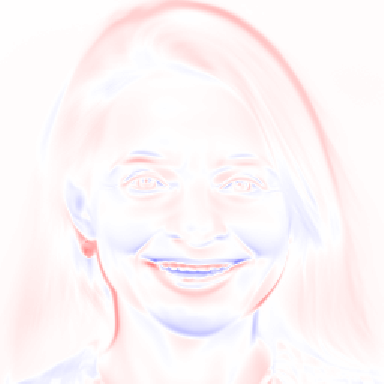} & 
\includegraphics[height=1.7cm]{figures/qual_res/multitarget/colorbar.svg.eps} \\
\end{tabular}

\caption{
\textbf{Granular reference values.} Counterfactual explanations generated by AC-RE and Diff-AE-RE for different reference values $\yref \in \{10,20,40,60,80\}$ for an image with an initial predicted age of 20. For each algorithm, we show the reconstructed image $\hat{\mathbf x}$ with corresponding counterfactuals input in the first row and input image $\mathbf x$ and heatmaps visualizing pixel-wise differences in the second row. AC-RE produces sparse, subtle changes. In contrast, Diff-AE-RE exhibits more intuitive but less sparse modifications for the lower age. This approach highlights key facial transformations associated with aging as well as a spurious feature}
\label{fig:exp_multitarget}
\end{figure}

Figure \ref{fig:exp_multitarget} presents the results of AC-RE and Diff-AE-RE for this experiment, including the reconstructed images and the counterfactuals for each reference value for the two algorithms. Underneath the explanations, we visualize the difference between the reconstruction and its counterfactuals by
plotting a heatmap showing the mean pixel value differences across color channels. Red indicates an increase in brightness, while blue indicates a decrease in brightness.
For this particular example, the initial prediction of the person is 20 years. This does not result in changes in the counterfactual for Diff-AE-RE for $\yref = 20$. However, AC-RE performs the refinement step regardless of the predicted value, always producing minor changes in the image.

We first analyze the explanations for AC-RE. As this algorithm produces sparse and subtle changes, the colors of the heatmap are naturally faint. As shown in Section \ref{sec:qual_res}, texture changes primarily influence prediction values. The algorithm mostly changes the area around the eyes and eyebrows to reach the reference values $\yref =10$ and $\yref =40$ from age 20. For the younger age, it smoothens the skin, while for the older age, it darkens the skin and adds texture. The most apparent changes with increasing age to $\yref = 60$ and $\yref = 80$ occur around the eyes the intensity of the smile lines, and also the darkening of the skin. Moreover, we observe one of the same key features of the predictor as observed in the original ACE paper \cite{jeanneretAdversarialCounterfactualVisual2023}: The cheek color gets redder with increasing reference value.

Diff-AE-RE in contrast produces heatmaps that are much more intuitively interpretable. We first observe that for $\yref = 10$, the regressor favors darker, rounder eyes and darker hair and skin. In addition, the neck seems to get slimmer. Noticeably, the whole image appears to be darker, which shows the problem of maintaining sparsity for this method. However, changes appear to be sparser for the latter reference values. 

The transition from $\yref=20$ to $\yref=40$ strongly suggests the presence of a spurious feature, as the main modification is the addition of an earring, consistent with the findings in Section \ref{sec:qual_res}. Beyond this, the strongest changes appear to be in the shape of the smile and nose, along with minor changes to the lips and the width of the neck. As age progresses to $\yref = 60$, these features become even more pronounced. In addition, the jaw and neck shape gain importance, as well as the decrease in hair volume and the number of wrinkles. For $\yref=80$, further modifications include thinning of the lips and eyebrows, complete disappearance of the second row of teeth, rounder chin, and even wider neck. The explanations provided by Diff-AE-RE reflect a combination of natural progressive aging processes and potential dataset biases.

\subsection{Ablation Study}\label{sec:ablation}

% \begin{itemize}
%     \item Choice and strength of distance function and pixel vs latent space application
%     \item See Figure \ref{fig:ablation_dist} for distance ablation example. No distance creates very strong changes in the images. l2 seems to limit but does not produce as much semantic changes as l1 distance.
%     \item Choice of Optimization algorithm: SGD and SGD + Momentum can barely move the CE in the latent space. Need to choose a very low learning rate costing computation time. When choosing Adam we don't have that problem.
%     \item Learning rate: by choosing a lower learning rate we achieve smoother CE results. Higher learning rate decreases computation time but comes with unrecoverable risk of leaving the data manifold, producing terrible images.
% \end{itemize}

To better understand the contribution of each component of Diff-AE-RE, we conduct an ablation study. For an applicable study on AC-RE, we refer to the ablation study on ACE \cite{jeanneretAdversarialCounterfactualVisual2023}.

\begin{figure}
\centering
% \begin{subfigure}{\textwidth} % Full width for the first row
    \centering
    \begin{tabular}{cccc} % Two columns, no extra space
    $\mathbf x$ & No Distance & $\ell_1(\zsem, \zsem')$ & $\ell_2(\zsem, \zsem')$ \\
    
    \includegraphics[width=0.24\textwidth]{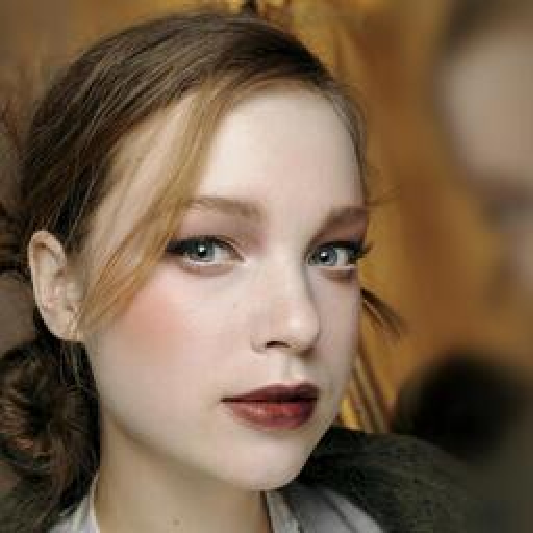} &
    \includegraphics[width=0.24\textwidth]{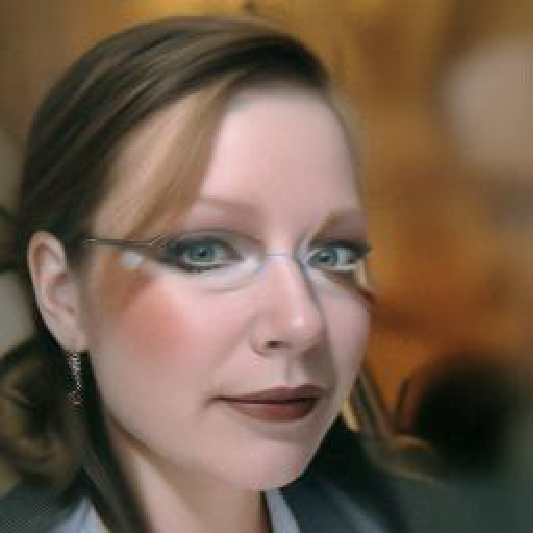} &
    \includegraphics[width=0.24\textwidth]{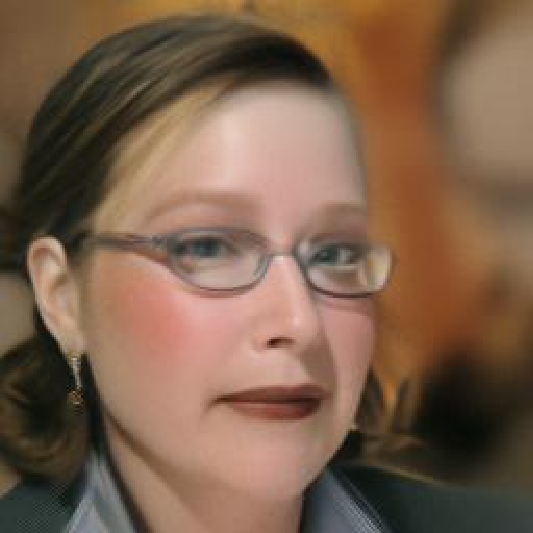} &
    \includegraphics[width=0.24\textwidth]{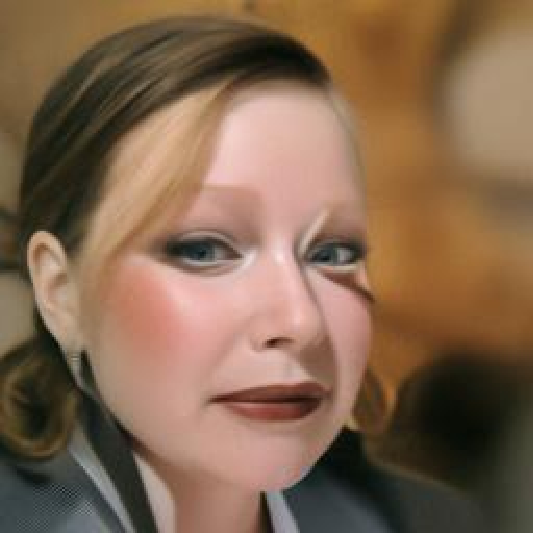} \\
    \end{tabular}
% \end{subfigure}
\caption{
    \textbf{Effect of distance function in the latent space with $\lambda_d = 10^{-5}$} We show the input image $\mathbf x$ and the generated counterfactuals with either no, $\ell_1$- and $\ell_2$-distance for the latent codes of the input $\zsem$ and counterfactual $\zsem'$. Without a distance function, alterations are aggressive, with some non-semantic changes. While the $\ell_2$-distance somewhat limits this, $\ell_1$-distance limits changes to be semantically meaningful while increasing the quality of the explanation
}
\label{fig:ablation_dist}
\end{figure}

First, we explore the choice and distance function and its regularization constant $\lambda_d$. For this we empirically tested combinations of the $
\ell_1$ and $\ell_2$ distance functions in the pixel and latent space, as well as a range of regularization constants $10^{-6} \leq \lambda_d \leq 10^{-2}$. We find that using no distance constraint resulted in aggressive image changes with some non-semantic changes. Applying the distance function on the pixel space results in counterfactuals with minimal changes in the image and the predicted age. 
Although the $\ell_2$ distance limited the alterations somewhat, applying the $\ell_1$ distance on the latent space produces more semantically meaningful and higher-quality counterfactuals. Regarding the regularization constant,
we find that choosing $\lambda_d = 10^{-5}$ strikes a balance between generating expressive changes and preventing domination of the loss. We show examples of these effects with $\lambda_d =10^{-5}$  in Figure \ref{fig:ablation_dist}.

Second, we evaluated the choice of the optimization algorithm. Stochastic Gradient Descent (SGD) and SGD with momentum struggled to effectively navigate the latent space, requiring impractically low learning rates. In contrast, Adam \cite{kingmaAdamMethodStochastic2017} efficiently optimized towards the reference value. Finally, we examine the learning rate, observing that lower rates yielded smoother and more realistic counterfactuals, while higher rates risked exiting the data manifold, leading to poor image quality. This leads us to empirically choose a learning rate of $0.002$ for Adam.

% 5. Conclusion
\section{Conclusion}\label{sec:conclusion}
% ---------------- CONTENT ----------------
% Conclusion
%    - Summary of key findings
%    - Strengths of the proposed method
%    - Future work directions
% ---------------- CONTENT ----------------
In this paper, we addressed the underexplored area of counterfactual explanations for image regression tasks by proposing two novel methods utilizing diffusion-based generative models: Adversarial Counterfactual Regression Explanations (AC-RE) operating in pixel space and Diffusion Autoencoder Regression Explanations (Diff-AE-RE) operating in latent space. Our methods perform similarly to classification-based counterparts and successfully generate realistic, semantic, and smooth counterfactuals. They provide valuable insights into the decision-making process of regression models and reveal spurious correlations. We demonstrated that feature changes in regression counterfactuals are dependent on the prediction region, with larger semantic alterations required for significant value shifts. Furthermore, we observed a trade-off between sparsity and quality, with AC-RE offering greater sparsity and Diff-AE-RE providing higher quality and semantic flexibility. This research paves the way for a better understanding of image regression models. Future work can apply these techniques to state-of-the-art diffusion models \cite{rombachHighResolutionImageSynthesis2022} and examine the theoretical implications of the Diffusion Autoencoder for Diffeomorphic Counterfactuals \cite{dombrowskiDiffeomorphicCounterfactualsGenerative2024}.

% ---------------- MAIN CONTENT ----------------

\begin{credits}
\subsubsection{\ackname} 
This work was supported by \textbf{BASLEARN}---TU Berlin/BASF Joint Laboratory, co-financed by TU Berlin and BASF SE.

\subsubsection{\discintname}
% It is now necessary to declare any competing interests or to specifically
% state that the authors have no competing interests. Please place the
% statement with a bold run-in heading in small font size beneath the
% (optional) acknowledgments\footnote{If EquinOCS, our proceedings submission
% system, is used, then the disclaimer can be provided directly in the system.},
% for example: 
The authors have no competing interests to declare that are
relevant to the content of this article. 
% Or: Author A has received research
% grants from Company W. Author B has received a speaker honorarium from
% Company X and owns stock in Company Y. Author C is a member of committee Z.
\end{credits}
\bibliographystyle{splncs04}
% \bibliography{references.bib, old.bib}
\bibliography{paper.bib}




% LNCS Homepage, \url{http://www.springer.com/lncs}, last accessed 2023/10/25
% \end{thebibliography}

% ---------------- APPENDIX ----------------
% \input{chapters/6_appendix}
\end{document}